\documentclass[11pt,a4paper]{article}
\usepackage{acl2015}
\usepackage{times}
\usepackage{url}
\usepackage{latexsym}
\usepackage{graphicx}
\usepackage{algorithm}
\usepackage{algorithmic}
 \usepackage{verbatim}
\usepackage{amsfonts}

\title{Semantic Relation Classification via Convolutional Neural Networks with Simple Negative Sampling}

\author{Kun Xu$^{1}$, Yansong Feng$^{1}$, Songfang Huang$^{2}$, Dongyan Zhao$^{1}$\\
$^{1}$Institute of Computer Science \& Technology, Peking University, Beijing, China\\
$^{2}$IBM China Research Lab, Beijing, China \\
\{xukun, fengyansong, zhaodongyan\}@pku.edu.cn,  huangsf@cn.ibm.com\\
}

\date{}

\begin{document}
\maketitle
\begin{abstract}
Syntactic features play an essential role in identifying relationship in a sentence. Previous neural network models 
often suffer from irrelevant information introduced
when subjects and objects are in a long distance.  
In this paper, we propose to learn more robust relation representations from the shortest dependency path through a 
convolution neural network. We further 
propose
a straightforward negative sampling strategy to improve the assignment of subjects and objects.  
Experimental results show that our method outperforms the state-of-the-art methods on the SemEval-2010 Task 8 dataset.
\end{abstract}

\section{Introduction}
The relation extraction (RE) task can be defined as follows: given a sentence $S$ with a pair of nominals $e_1$ and $e_2$, we aim to identify the relationship between $e_1$ and $e_2$. RE is typically investigated in a classification style, where many features have been proposed, e.g., \newcite{hendrickx} 
designed 16 types of features including POS, WordNet, FrameNet, dependency parse features, etc. Among them,  syntactic features are considered to bring significant improvements in extraction accuracy \cite{DBLP:conf/naacl/BunescuM05}. Earlier attempts to encode syntactic information are mainly kernel-based methods, such as the convolution tree kernel \cite{DBLP:conf/coling/QianZKZQ08}, subsequence kernel \cite{DBLP:conf/nips/BunescuM05}, and dependency tree kernel \cite{DBLP:conf/naacl/BunescuM05}. 

With the recent success of neural networks in NLP, different neural network models are proposed to learn syntactic features from raw sequences of words or constituent parse trees\cite{zeng-EtAl:2014:Coling,DBLP:conf/emnlp/SocherHMN12}, which have been proved effective, but, often suffer from irrelevant subsequences or clauses, especially when subjects and objects are in a longer distance. For example, in the sentence, ``The [singer]$_{e1}$ , who performed three of the nominated songs, also caused a [commotion]$_{e2}$ on the red carpet'', the \textit{who} clause is used to modify subject $e_1$, but is unrelated to the \textit{Cause-Effect} relationship between \textit{singer} and \textit{commotion}. Incorporating such 
information into the model will hurt the extraction performance. We therefore propose to learn a more robust relation representation from a convolution neural network model that works on the simple dependency path between subjects and objects, which naturally characterizes the relationship between two nominals and avoids negative effects from other irrelevant chunks or clauses.



Our second contribution is the introduction of a negative sampling strategy into the CNN models to address the relation directionality, i.e., properly assigning the subject and object within a relationship. In the above \textit{singer} example, (\textit{singer},  \textit{commotion}) hold the \textit{Cause-Effect} relation, while (\textit{commotion}, \textit{singer}) not. Previous works do not fully investigate the differences between subjects and objects in the utterance, and simply transform  a (\textit{K}+1)-relation task into a (2$\times$\textit{K}+1) classification task, where 1 is the \textit{other} relation. Interestingly, we find that dependency paths naturally offer the relative positions of subjects and objects through the path directions.  
In this paper, we propose to model the relation directionality by exploiting the dependency path to learn the assignments of subjects and objects using a straightforward negative sampling method, which adopts the shortest dependency path from the object to the subject as a negative sample. Experimental results show that the negative sampling method significantly improves the performance, and our model outperforms the-state-of-the-art methods on the SemEval-2010 Task 8 dataset.

\begin{figure*}[htp!]
\centering\includegraphics[width=0.9\textwidth]{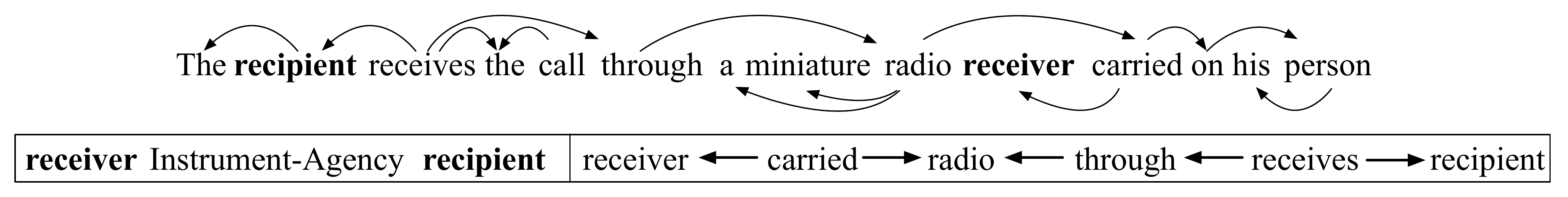} 
\caption{ \label{dep} The shortest dependency path representation 
for an example sentence from SemEval-08.}
\end{figure*}
\section{The Shortest Path Hypothesis}
If $e_1$ and $e_2$ are two nominals mentioned in the same sentence, we assume that the shortest path between $e_1$ and $e_2$ describes their relationship. This is because 
(1) if $e_1$ and $e_2$ are arguments of the same predicate, then their shortest path should
 pass through that predicate; 
(2) if $e_1$ and $e_2$ belong to different predicate-argument structures, their shortest path will pass through a sequence of predicates,  
and any consecutive predicates  
will share a common argument. Note that, the order of the predicates 
on the path indicates the proper assignments of subjects and objects for that relation. For example, in Figure~\ref{dep}, the dependency path consecutively passes through  \textit{carried} and \textit{receives}, which together implies that in the \textit{Instrument-Agency} relation, the subject and object play a sender and receiver role, respectively. 


\begin{figure}[htp!]
\centering\includegraphics[width=0.5\textwidth]{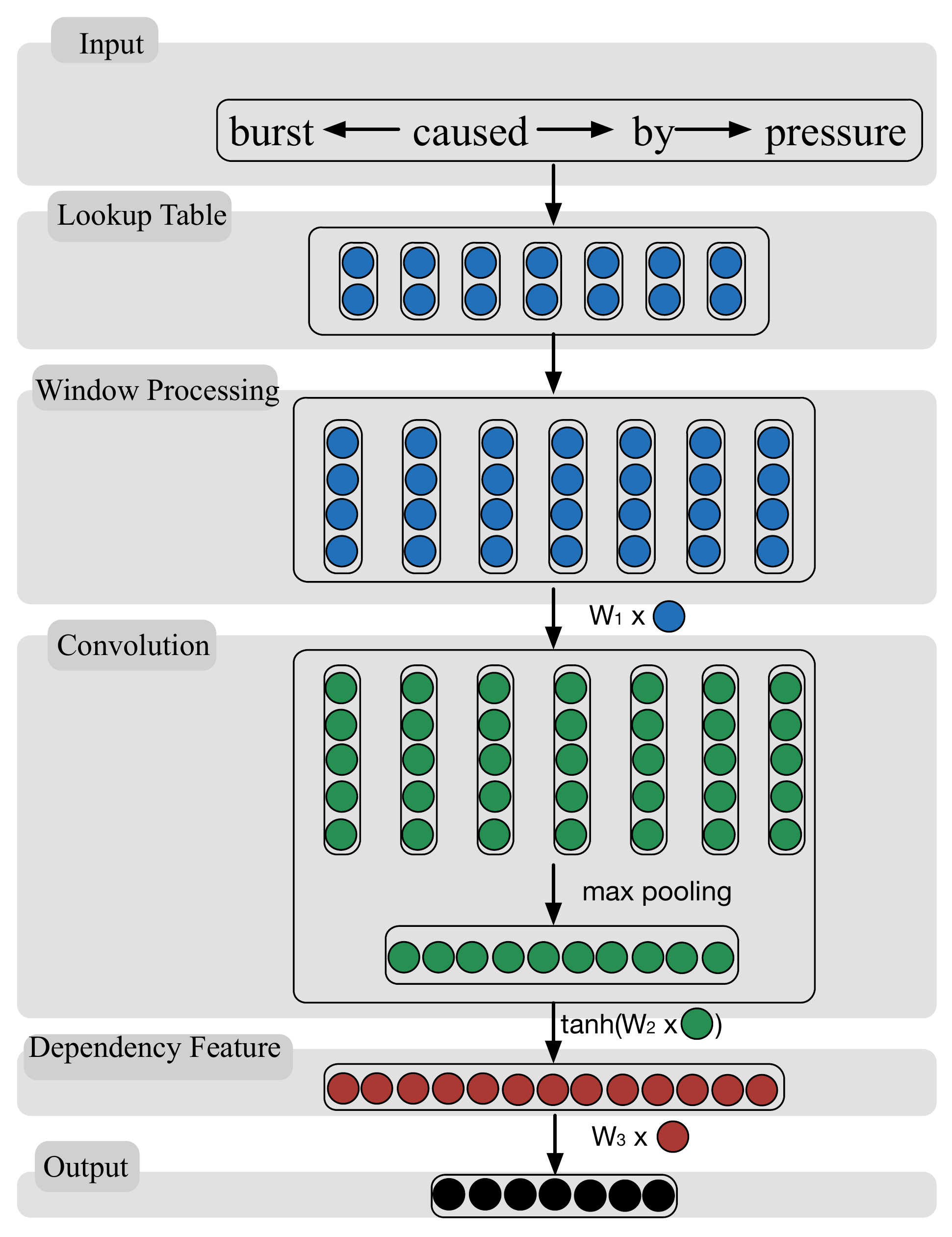} 
\caption{ \label{cnn} Architecture of the convolution neural network.}
\end{figure}
\section{A Convolutional Neural Network Model}
%

Our model successively takes the shortest dependency path (i.e, the words, dependency edge directions, and dependency labels) from the subject  to the object  as input, passes it through the lookup table layer, produces local features around each node on the dependency path, and combines these features into a global feature vector that are then fed to a softmax classifier. Each dimension of the output vector
indicates the confidence score of the corresponding relation.

In the \textit{lookup table} step, each node (i.e. word, label or arrow) in the dependency path is transformed into a vector by looking up the embedding matrix $W_e \in \mathbb{R}^{d \times |\mathbb{V}|}$, where $d$ is the dimension of a vector and $\mathbb{V}$ is a set of all nodes we consider.

\paragraph{Convolution}
To capture the local features around each node of the dependency path, we consider a fixed size window of nodes around each node in the \textit{window processing} component, producing a matrix of node features of fixed size $d_{w} \times 1$, where $d_{w} = d \times w$ and $w$ is the window size. This matrix can be 
built by concatenating the vectors of nodes within the window. 

In the convolutional layer, we use a linear transformation  $W_1 \in \mathbb{R}^{n_1\times d_w}$ to extract local features around each window of the given sequence, where $n_1$ is the size of hidden layer 1. 
The resulting matrix  $Z$ has size of $n_1\times t$, where $t$ is the number of nodes in the input dependency path.

We can see that $Z$ captures local contextual information in the dependency path. 
Therefore, we perform a max pooling over  $Z$ to produce a global feature vector in order to capture the 
most useful local features produced by the convolutional layer  \cite{DBLP:journals/jmlr/CollobertWBKKK11}, which has a fixed size of $n_1$, 
 independent of the dependency path length.
%

\paragraph{Dependency based Relation Representation}
To extract more meaningful features, we choose hyperbolic tanh as the non-linearity function in the second hidden layer, which has the advantage of being slightly cheaper to compute, while leaving the generalization performance unchanged. $W_2 \in \mathbb{R}^{n_2\times n_1}$ is the linear transformation matrix, where $n_2$ is the size of hidden layer 2. The output vector can be considered as higher level syntactic features, which is then fed to a softmax classifier.

\paragraph{Objective Function and Learning}
The softmax classifier is used to predict a $K$-class distribution $d(x)$, where  $K$ is the size of all possible relation types, and 
the transformation matrix is $W_3 \in \mathbb{R}^{K\times n_2}$. 
We denote $t(x) \in \mathbb{R}^{K\times 1}$ as the target distribution vector\footnote{Note that, there may be more than one relation existing between two nominals. A dependency path thus may correspond to multiple relations.}:
the entry $t_{k}(x)$ is the probability that the dependency path describes the \textit{k}-th relation. 
We compute the cross entropy error between $t(x)$ and $d(x)$, and further define the objective function over all training data:
\begin{displaymath}
J(\theta) = - \sum_{x} \sum_{k=1}^{K} t_k(x) \log d_k(x) + \lambda||\theta||^{2} 
\end{displaymath}
where $\theta = (W_e, W_1, W_2, W_3)$ is the set of model parameters to be learned, and 
$\lambda$ is a vector of regularization parameters. The model parameters $\theta$ can be efficiently computed via 
backpropagation through network structures. To minimize $J(\theta)$, we apply stochastic gradient descent (SGD) with AdaGrad \cite{DBLP:journals/jmlr/DuchiHS11} in our experiments\footnote{We omit detailed formulas for the limitation of space.}.

\begin{table}
\small
\centering
\begin{tabular}{c|c|c}
\hline
Train Strategy & Test Strategy & F1(\%) \\
\hline
Blind & Blind & 79.3\\
\hline
Sighted & Blind & 81.3\\
\hline
Sighted & Sighted & 89.2\\
\hline
\end{tabular}
\caption{Performances on the development set with different train and testing strategies.}
\label{tab:ns_results}
\end{table}
\section{Negative Sampling}
\label{sec:ns}
We start by presenting three pilot experiments on the development set. In the first one, we assume that the assignment of the subject and object for a relation is not given (blind), we simply extract features from $e_1$ to $e_2$, and test it in a blind setting as well. In the second one, we assume that the assignment is given (sighted) during training, but still blind in the test phase. The last one is assumed to give the assignment during both  training and test steps. The results are listed in Table~\ref{tab:ns_results}.

The third experiment can be seen as an upper bound, where we do not need to worry about the assignments 
of subjects and objects. By comparing the first and the second one, we can see that when adding 
assignment information during training, our model can be significantly improved, indicating that 
our dependency based representation can be used to  learn the assignments of subjects/objects, and injecting
 better understandings of such assignments during training is crucial to the performance. 
We admit that models with more complex structures can better handle these considerations. 
However, we find that this can be achieved by simply feeding typical negative samples to the model 
and let the model learn from such negative examples to correctly choose the right assignments of subjects and objects. 
In practice, we can treat the opposite assignments of subjects and the objects as negative examples. 
Note that, the dependency path of the wrong assignment is different from that of the correct assignment, which essentially offers the information for the model to learn to distinguish the subject and the object.

\section{Experimental Evaluation}
We evaluate our model on the SemEval-2010 Task 8 \cite{hendrickx}, which contains 10,717 annotated examples, including 8,000 instances for training and 2,717 for test. We randomly sampled 2,182 samples from the training data for validation.

Given a sentence, we first find the shortest dependency path connecting two marked nominals, resulting in two dependency paths corresponding to two opposite subject/object directions, and then make predictions for the two paths, respectively. 
We choose the relation \textit{other} \textbf{if and only if} both predictions are \textit{other}. And for the rest cases, 
we choose the non-\textit{other} relation with highest confidence as the output, since ideally, for a non-\textit{other} instance, 
our model will output the correct label for the right  subject/object direction and an   \textit{other}  label for the wrong direction.
We evaluate our models by macro-averaged F1 using the official evaluation script.

%
%

We initialized $W_e$ with 50-dimensional word vectors trained by \newcite{DBLP:conf/acl/TurianRB10}.
We tuned the hyper parameters using the development set for each experimental setting. The hyper parameters include $w$, $n_1$, $n_2$, and regularization parameters for $W_e$, $W_1$, $W_2$ and $W_3$. The best setting was obtained with the values: 3, 200, 100, $10^{-4}$, $10^{-3}$, $10^{-4}$ and $2\times 10^{-3}$, respectively.

\paragraph{Results and Discussion}
\begin{table}
\small
\centering
\begin{tabular}{l|l|c}
\hline
Method & Feature Sets & F1 \\
\hline
\hline
SVM & 16 types of features & 82.2 \\
\hline
RNN & - & 74.8 \\
& +POS, NER, WordNet & 77.6 \\
\hline
MVRNN & - & 79.1 \\
 & +POS, NER, WordNet & 82.4 \\
\hline
CNN & - & 78.9 \\
{\tiny \cite{zeng-EtAl:2014:Coling}}& +WordNet,words around nominals & 82.7 \\\hline
depCNN & - & 81.3\\
depLCNN & - & 81.9\\
depLCNN & +WordNet,words around nominals & 83.7 \\
depLCNN+NS & - & \textbf{84.0} \\
& +WordNet,words around nominals & \textbf{85.6} \\
\hline
\end{tabular}
\caption{Comparisons of our models with other methods on the SemEval 2010 task 8.}
\label{tab:results}
\end{table}

\begin{table}
\small
\centering
\begin{tabular}{l|c}
\hline
Negative sampling schemes & F1 \\
\hline
No negative examples & 81.3\\
\hline
Randomly sampled negative examples from NYT &  83.5\\
\hline
Dependency paths from the object to subject &  \textbf{85.4}\\
\hline
\end{tabular}
\caption{Comparisons of different negtive sampling methods on the development set.}
\label{tab:results_2}
\end{table}

Table~\ref{tab:results} summarizes  
 the performances of our model, depLCNN+NS(+), and state-of-the-art models, SVM\cite{hendrickx}, 
RNN, MV-RNN\cite{DBLP:conf/emnlp/SocherHMN12}, and CNN\cite{zeng-EtAl:2014:Coling}.
For fair comparisons, we also add two types of lexical features,  
WordNet hypernyms and words around nominals, as part of input vector to the final \texttt{softmax} layer.

We can see that  our vanilla depLCNN+NS, without extra lexical features, still outperforms, by a large margin, previously 
reported best systems, MVRNN+ and CNN+, both of which have taken extra lexical features into account, showing
that our treatment to dependency path can learn a robust and effective relation representation.  
 When augmented with similar lexical features, our depLCNN+NS further improves by 1.6\%, significantly better than 
any other systems.

Let us first see the comparisons among plain versions of depLCNN (taking both dependency directions and labels into account),
 depCNN (considering the directions of dependency edges only), MVRNN and CNN, which all work in a 2$\times$\textit{K}+1 fashion.
We can see that the both of our depCNN and depLCNN outperforms MVRNN and CNN by at least 2.2\%, 
indicating that our treatment  is better than previous conventions in capturing syntactic structures for relation extraction.
And note that depLCNN, with extra considerations for dependency labels, performs even better than depCNN,  
showing that dependency labels offer more discriminative information that benefits  the relation extraction task.

%

And when we compare plain depLCNN and depLCNN+NS (without lexical features), 
we can see that our Negative Sampling strategy brings an improvement of 2.1\% in F1. When both of the two models
are augmented with extra lexical features, our NS strategy still gives an improvement of 1.9\%.
 These comparisons further show that  our NS strategy can drive our model to learn proper assignments of subjects/objects 
for a relation. 

Next, we will have a close look at the effect of our Negative Sampling method.
We conduct additional experiments on the development set to compare two different negative sampling methods. As a baseline, we randomly sampled 
8,000 negative examples from the NYT dataset \cite{DBLP:conf/acl/ChenFHQZ14}. 
For our proposed NS, we create a negative example from each \textit{non-other} instance in the training set, 
6,586 in total. 
As shown in Table~\ref{tab:results}, it is no doubt that introducing more negative examples improves the performances.
 We can see that our model still benefits from the randomly sampled negative examples, which may help our model learn to refine the margin between the positive and negative examples. 
However, with similar amount of negative examples, treating the reversed dependency paths from objects to subjects as negative examples can achieve a better performance (85.4\% F1), improving random samples by 1.9\%. This again proves that dependency paths provide useful clues  to 
reveal the assignments of subjects and objects, and a model can learn  from such reversed paths as negative examples to make correct assignments.
%
Beyond the relation extraction task, we believed the proposed Negative Sampling method 
has the potential to benefit other NLP tasks, which we leave for future work.

\section{Conclusion}

In this paper, we exploit a convolution neural network to learn more robust and effective relation representations from shortest dependency paths 
for relation extraction. We further propose a simple  negative sampling method to help make correct assignments for subjects and objects within a relationship. 
Experimental results show that our model significantly outperforms state-of-the-art systems and our treatment to dependency paths
 can well capture the syntactic features for relation extraction. 


\bibliographystyle{acl}
\bibliography{acl2015}

\begin{thebibliography}{}

\bibitem[\protect\citename{Bunescu and
  Mooney}2005a]{DBLP:conf/naacl/BunescuM05}
Razvan~C. Bunescu and Raymond~J. Mooney.
\newblock 2005a.
\newblock A shortest path dependency kernel for relation extraction.
\newblock In {\em {HLT/EMNLP} 2005, Human Language Technology Conference and
  Conference on Empirical Methods in Natural Language Processing, Proceedings
  of the Conference, 6-8 October 2005, Vancouver, British Columbia, Canada}.

\bibitem[\protect\citename{Bunescu and Mooney}2005b]{DBLP:conf/nips/BunescuM05}
Razvan~C. Bunescu and Raymond~J. Mooney.
\newblock 2005b.
\newblock Subsequence kernels for relation extraction.
\newblock In {\em Advances in Neural Information Processing Systems 18 [Neural
  Information Processing Systems, {NIPS} 2005, December 5-8, 2005, Vancouver,
  British Columbia, Canada]}, pages 171--178.

\bibitem[\protect\citename{Chen \bgroup et al.\egroup
  }2014]{DBLP:conf/acl/ChenFHQZ14}
Liwei Chen, Yansong Feng, Songfang Huang, Yong Qin, and Dongyan Zhao.
\newblock 2014.
\newblock Encoding relation requirements for relation extraction via joint
  inference.
\newblock In {\em Proceedings of the 52nd Annual Meeting of the Association for
  Computational Linguistics, {ACL} 2014, June 22-27, 2014, Baltimore, MD, USA,
  Volume 1: Long Papers}, pages 818--827.

\bibitem[\protect\citename{Collobert \bgroup et al.\egroup
  }2011]{DBLP:journals/jmlr/CollobertWBKKK11}
Ronan Collobert, Jason Weston, L{\'{e}}on Bottou, Michael Karlen, Koray
  Kavukcuoglu, and Pavel~P. Kuksa.
\newblock 2011.
\newblock Natural language processing (almost) from scratch.
\newblock {\em Journal of Machine Learning Research}, 12:2493--2537.

\bibitem[\protect\citename{Duchi \bgroup et al.\egroup
  }2011]{DBLP:journals/jmlr/DuchiHS11}
John~C. Duchi, Elad Hazan, and Yoram Singer.
\newblock 2011.
\newblock Adaptive subgradient methods for online learning and stochastic
  optimization.
\newblock {\em Journal of Machine Learning Research}, 12:2121--2159.

\bibitem[\protect\citename{Hendrickx \bgroup et al.\egroup }2010]{hendrickx}
Iris Hendrickx, Su~Nam Kim, Zornitsa Kozareva, Preslav Nakov, Diarmuid~\'O
  S\'eaghdha, Sebastian Pad\'o, Marco Pennacchiotti, Lorenza Romano, and Stan
  Szpakowicz.
\newblock 2010.
\newblock Semeval-2010 task 8: Multi-way classification of semantic relations
  between pairs of nominals.
\newblock In {\em Proceedings of SemEval-2}, Uppsala, Sweden.

\bibitem[\protect\citename{Qian \bgroup et al.\egroup
  }2008]{DBLP:conf/coling/QianZKZQ08}
Longhua Qian, Guodong Zhou, Fang Kong, Qiaoming Zhu, and Peide Qian.
\newblock 2008.
\newblock Exploiting constituent dependencies for tree kernel-based semantic
  relation extraction.
\newblock In {\em {COLING} 2008, 22nd International Conference on Computational
  Linguistics, Proceedings of the Conference, 18-22 August 2008, Manchester,
  {UK}}, pages 697--704.

\bibitem[\protect\citename{Socher \bgroup et al.\egroup
  }2012]{DBLP:conf/emnlp/SocherHMN12}
Richard Socher, Brody Huval, Christopher~D. Manning, and Andrew~Y. Ng.
\newblock 2012.
\newblock Semantic compositionality through recursive matrix-vector spaces.
\newblock In {\em Proceedings of the 2012 Joint Conference on Empirical Methods
  in Natural Language Processing and Computational Natural Language Learning,
  EMNLP-CoNLL 2012, July 12-14, 2012, Jeju Island, Korea}, pages 1201--1211.

\bibitem[\protect\citename{Turian \bgroup et al.\egroup
  }2010]{DBLP:conf/acl/TurianRB10}
Joseph~P. Turian, Lev{-}Arie Ratinov, and Yoshua Bengio.
\newblock 2010.
\newblock Word representations: {A} simple and general method for
  semi-supervised learning.
\newblock In {\em {ACL} 2010, Proceedings of the 48th Annual Meeting of the
  Association for Computational Linguistics, July 11-16, 2010, Uppsala,
  Sweden}, pages 384--394.

\bibitem[\protect\citename{Zeng \bgroup et al.\egroup
  }2014]{zeng-EtAl:2014:Coling}
Daojian Zeng, Kang Liu, Siwei Lai, Guangyou Zhou, and Jun Zhao.
\newblock 2014.
\newblock Relation classification via convolutional deep neural network.
\newblock In {\em Proceedings of COLING 2014, the 25th International Conference
  on Computational Linguistics: Technical Papers}, pages 2335--2344, Dublin,
  Ireland, August. Dublin City University and Association for Computational
  Linguistics.

\end{thebibliography}

\end{document}